\documentclass[letterpaper, 10 pt, conference]{ieeeconf}  

\IEEEoverridecommandlockouts                              

\usepackage{amsmath} 
\usepackage{amssymb}  
\usepackage{cite}
\usepackage{hyperref}   
\hypersetup{pdfstartview=FitH,
            colorlinks=true,
            linkcolor=red,
            anchorcolor=blue,
            citecolor=green
            }
\usepackage[T1]{fontenc}
\usepackage{pifont}
\usepackage{graphicx}
\usepackage{makecell}   
\usepackage[table]{xcolor}
\usepackage{multirow}
\usepackage{makecell}
\usepackage{booktabs}
\usepackage{subfigure}

\definecolor{defaultcolor}{gray}{.9}
\newcommand{\default}[1]{\cellcolor{defaultcolor}{#1}}
\newcommand{\tabincell}[2]{\begin{tabular}{@{}#1@{}}#2\end{tabular}}

\title{\LARGE \bf
DMTrack: Spatio-Temporal Multimodal Tracking via Dual-Adapter
}

\author{
Weihong Li$^{1, 2}$,
Shaohua Dong$^{3}$,
Haonan Lu$^{4}$,
Yanhao Zhang$^{4}$,
Heng Fan$^{3,\dagger}$,
Libo Zhang$^{1, 2,\dagger,*}$%
\thanks{$^{\dagger}$Equal advising and co-last authors.}
\thanks{$^*$Corresponding author: libo@iscas.ac.cn}
\thanks{$^{1}$Hangzhou Institute for Advanced Study, University of Chinese Academy of Sciences}
\thanks{$^{2}$Institute of Software, Chinese Academy of Sciences}
\thanks{$^{3}$Department of Computer Science and Engineering, University of North Texas}
\thanks{$^{4}$OPPO AI Center}
}

\begin{document}

\maketitle
\thispagestyle{empty}
\pagestyle{empty}

\begin{abstract}

In this paper, we explore adapter tuning and introduce a novel dual-adapter architecture for spatio-temporal multimodal tracking, dubbed DMTrack. The key of our DMTrack lies in two simple yet effective modules, including a spatio-temporal modality adapter (STMA) and a progressive modality complementary adapter (PMCA) module. The former, applied to each modality alone, aims to adjust spatio-temporal features extracted from a frozen backbone by self-prompting, which to some extent can bridge the gap between different modalities and thus allows better cross-modality fusion. The latter seeks to facilitate cross-modality prompting progressively with two specially designed pixel-wise shallow and deep adapters. The shallow adapter employs shared parameters between the two modalities, aiming to bridge the information flow between the two modality branches, thereby laying the foundation for following modality fusion, while the deep adapter modulates the preliminarily fused information flow with pixel-wise inner-modal attention and further generates modality-aware prompts through pixel-wise inter-modal attention. With such designs, DMTrack achieves promising spatio-temporal multimodal tracking performance with merely \textbf{0.93M} trainable parameters. Extensive experiments on five benchmarks demonstrate that DMTrack achieves state-of-the-art results. Our code and models will be available at \href{https://github.com/Nightwatch-Fox11/DMTrack}{https://github.com/Nightwatch-Fox11/DMTrack}.

\end{abstract}

\section{INTRODUCTION}
\label{sec:intro}

Over the past decades, visual object tracking has played a vital role in computer vision. The remarkable surge of excellent tracking frameworks ~\cite{ostrack,artrackv2,odtrack,mcitrack,lin2022swintrack,lin2024tracking} has boosted numerous real-world applications~\cite{zhang2023dual,liu2012hand,itti2004automatic,xing2010multiple}. Despite the promising performance achieved by fine-tuning on large-scale benchmarks~\cite{vasttrack,fan2019lasot,huang2019got,muller2018trackingnet}, RGB-based object tracking still fails to handle “corner scenarios” under open-world settings, such as extreme illumination and occlusion of similar distractors. Therefore, multimodal tracking is emerging as a pivotal catalyst for advancing more robust tracking performance. 

\begin{figure}[t]
  \centering
  \setlength{\abovecaptionskip}{-1mm}
  \includegraphics[width=1\linewidth]{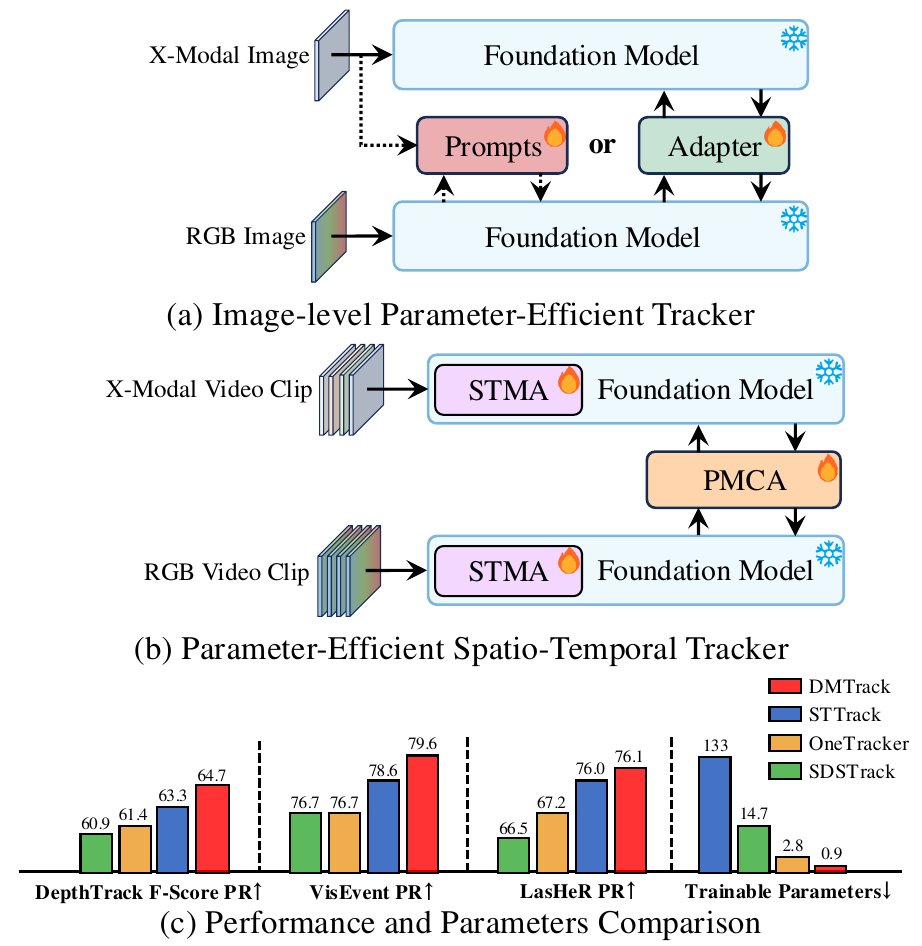}
  \caption{Comparison of existing unified multimodal trackers and our proposed DMTrack in frameworks (a)-(b) and performance (c). \emph{Best viewed in color for all figures in this paper.}}
  \vspace{-7mm}
  \label{fig:pipeline_compare}
\end{figure}

Due to the limited scale of downstream training data~\cite{li2021lasher,yan2021depthtrack,wang2023visevent}, dominant multimodal trackers typically leverage the power of foundation models pre-trained on RGB sequences. To handle this issue, researchers explore parameter-efficient training approaches for multimodal tracking. As demonstrated in Fig.~\ref{fig:pipeline_compare}~(a), by introducing only a few trainable parameters, some methods~\cite{protrack,vipt,bat} have pioneered the use of parameter-efficient fine-tuning (PEFT) techniques (\textit{e.g.}, prompt tuning, adapter tuning, \textit{etc.}) to adapt RGB-based foundational trackers for multimodal tracking tasks, sparking a trend of PEFT in this field. Recent efforts~\cite{untrack,onetracker} have further explored LoRA~\cite{lora} techniques in pursuit of unified multimodal tracking. However, these attempts still adopt an image-level tracking paradigm that relies on a fixed initial template frame and only model spatial relationships, thus limiting their ability to handle complicated situations with significant target appearance variations. 

Conversely, some trackers~\cite{mambavt,STTrack} begin to explore spatio-temporal multimodal tracking through fully fine-tuning on Mamba~\cite{mamba}-based architectures and incorporate global interaction between video streams from different modalities to jointly model spatio-temporal contexts. Although the incorporation of temporal information leads to performance gains, it also introduces a large number of trainable parameters and computational demands, resulting in high memory costs.


To mitigate these limitations, we propose a novel multimodal tracker, dubbed DMTrack, toward parameter-efficient spatio-temporal tracking. In contrast to existing non-temporal parameter-efficient multimodal trackers, we present the first attempt to extend PEFT to joint spatio-temporal context modeling. As shown in Fig.~\ref{fig:pipeline_compare}~(b), DMTrack freezes the entire foundation model and employs two separate branches to process different modalities. Each branch first performs pixel-wise inner-modality spatio-temporal modeling in a self-prompting manner, then progressively injects cross-modal complementary prompts, enriched with spatio-temporal cues, into the other modality branch on a per-pixel basis. All learned prompts are built upon the parameters of the foundation model.
Specifically, 1) For inner-modality spatio-temporal information incorporation, we adopt a simple template memory bank without temporal propagation to establish temporal relationships efficiently, and we design an STMA that enhances the spatio-temporal feature within the modality-specific template memory while simultaneously reducing the gap between modalities; 2) For inter-modality prompts generation, we propose a PCMA module that facilitates cross-modal interactions with linear complexity. The PCMA module features twin adapters: the shallow adapter establishes bidirectional cross-modal feature alignment via dense connections, while the deep adapter employs pixel-wise attention to refine fused representations and incorporate complementary modality guidance simultaneously.

We summarize our \textbf{contributions} as follows:

(1) We present DMTrack, a parameter-efficient framework that adapts pre-trained image-level RGB-based trackers for robust video-level multimodal tracking by integrating dual spatio-temporal adapter modules; (2) DMTrack performs cost-effective modeling of inner-modality spatio-temporal correlation and further reduces computational expenses by progressively generating cross-modal prompts on a pixel-wise basis; (3) To the best of our knowledge, we are the first to leverage adapters to explore spatio-temporal contextual modeling for multimodal tracking. By incorporating only 0.93M trainable parameters (accounting for 0.9\% of the total), DMTrack converges to optimal performance within a 5-hour training; (4) Extensive experiments demonstrate that DMTrack achieves state-of-the-art performance across five prevailing benchmark datasets, including DepthTrack, VOT-RGBD2022, VisEvent, LasHeR, and RGBT234.

\section{Related Works}
\label{sec:related_works}
\subsection{Multimodal Tracking}
Recent RGB-based tracking methods~\cite{mcitrack,artrackv2,odtrack} have achieved promising results on large-scale datasets~\cite{fan2019lasot,muller2018trackingnet,huang2019got}. However, despite the strong temporal mechanisms employed, single-modal tracking paradigms still struggle to tackle real-world challenges such as extreme illumination variations. As a result, multimodal trackers, which introduce auxiliary modalities to complement RGB, have gained significant attention. ViPT~\cite{vipt}, as an early method, injects auxiliary modalities cues into the RGB information stream with a prompt-tuning architecture. BAT~\cite{bat} introduces a bidirectional adapter that enables reciprocal interaction between the auxiliary modality and RGB. Although both methods leverage PEFT techniques to reduce training costs, they fail to account for the temporal information. MambaVT~\cite{mambavt} and STTrack~\cite{STTrack} jointly model spatio-temporal information by global interaction of video streams from different modalities with Mamba~\cite{mamba} architecture. Despite their reasonable performance, current spatio-temporal tracking methods rely on full fine-tuning strategies and global cross-modal interaction between video streams, thus suffering from prohibitive memory and computational demands. In this study, we pioneer a modality-specific adapter design for self-prompting spatio-temporal context in multimodal tracking. With such designs, we reduce the inherent gap between modalities for the following cross-modal prompts generation and avoid expensive global interactions among video tokens from two modalities.

\subsection{Parameter-Efficient Tuning} 
Different from full fine-tuning, PEFT has recently garnered significant attention due to its ability to substantially reduce the number of trainable parameters, offering an efficient approach to leverage pre-trained models. Originally developed for NLP~\cite{peft}, PEFT has since been adapted and applied to a variety of vision tasks~\cite{protrack,vipt,bat}. Some works~\cite{aim,stadapter,m2clip} begin to adapt large pre-trained image models (\textit{i.e.}, CLIP~\cite{clip}) for video downstream tasks. AIM~\cite{aim} proposed a joint spatio-temporal adaptation method to fine-tune pre-trained vision transformers. ST-Adapter~\cite{stadapter} introduced a parameter-efficient space-time adapter that effectively unleashes the power of CLIP for video understanding. Meanwhile, with the advent of ProTrack~\cite{protrack}, prompt-tuning was first applied to the tracking domain. Moreover, BAT and ViPT explore the potential of freezing the parameters of image-level trackers while incorporating various spatial adapters or prompts for multimodal tracking. Different from previous parameter-efficient trackers, we introduce spatio-temporal adapters to the multimodal tracking field for jointly modeling inner-modal spatio-temporal correlation, which to our knowledge has not been studied before. In addition to the STMA design, we incorporate pixel-wise attention mechanisms into the adapter architecture to generate modality-aware prompts for inter-modality interaction.

\subsection{Multimodal Fusion}
Multimodal fusion serves as a fundamental component in various perception tasks. In autonomous driving, existing transformer-based methods like TransFuser~\cite{transfuser} and TriTransNet~\cite{tritransnet} achieve cross-modal interaction through self-attention mechanisms, but they suffer from quadratic complexity that limits computational efficiency. Recent advances in efficient fusion strategies reveal two promising directions: TokenFusion~\cite{tokenfusion} enhances feature selectivity through dynamic token exchange between modalities, while GeminiFusion~\cite{geminifusion} introduces lightweight pixel-wise attention for multimodal semantic segmentation. Building upon these developments, we present a novel PMCA module that progressively integrates cross-modal complementary information through a twin adapter design. Our architecture features: a) A shallow bidirectional bridge adapter that synchronously aligns feature representations between modalities through shared dense connection layers, and b) A deep refinement adapter that employs a pixel-wise attention mechanism to modulate preliminary fused modality flow while iteratively injecting complementary guidance from the alternate modality. This dual-stage adaptation enables the progressive incorporation of cross-modal cues through parameter-efficient operations while preserving modality-specific characteristics

\begin{figure*}[!t]
	\centering
	\includegraphics[width=0.9\linewidth]{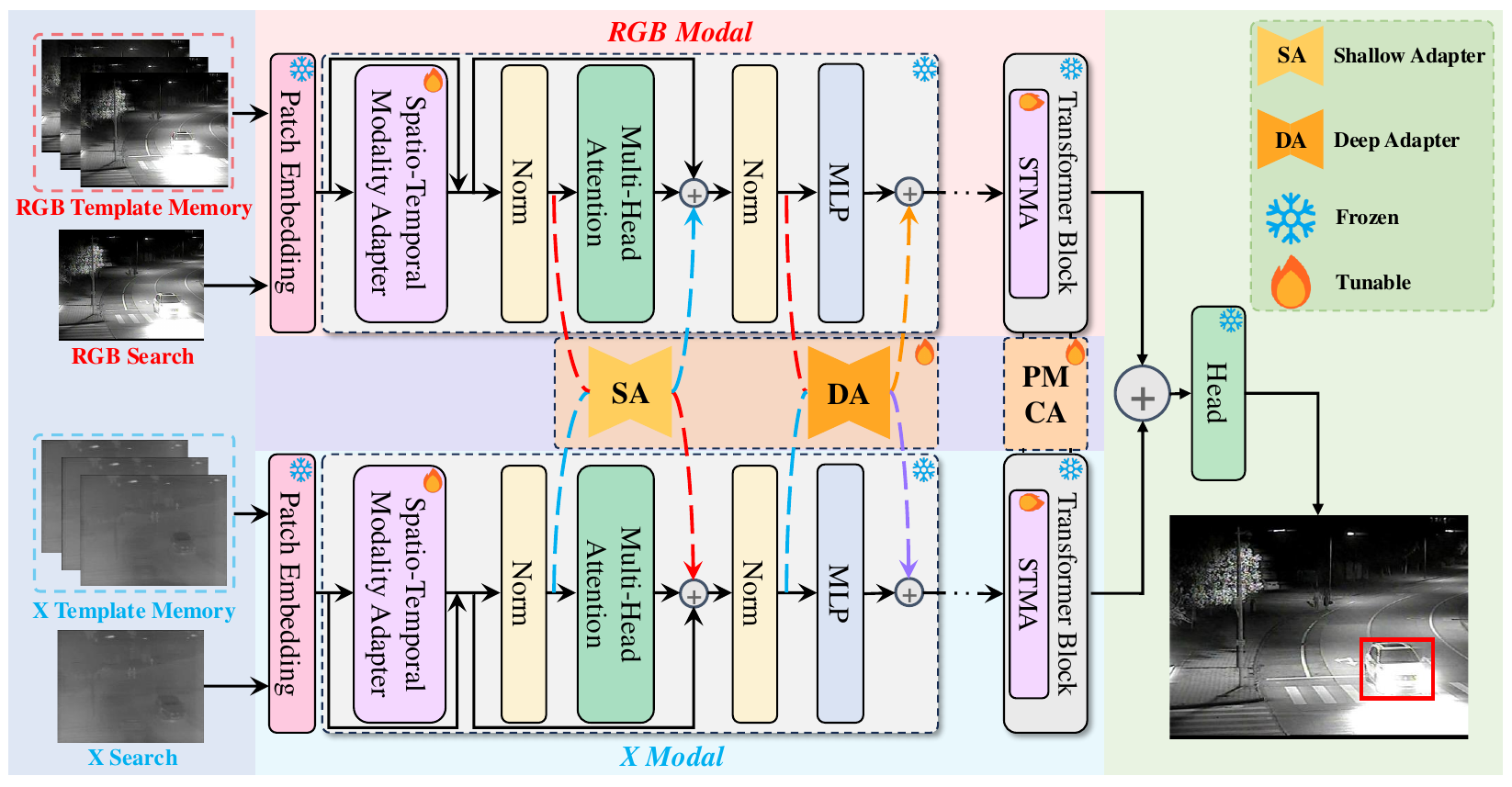}
	\caption{Overview of the proposed DMTrack. We first tokenize the template and search frames from each modality, then concatenate the resulting token sequences and process them through the frozen transformer architecture. Within each block structure, the STMA remains the only trainable component, specifically designed to produce self-prompts that encode intra-modal spatio-temporal relationships. The PMCA module bridges two processing branches through a twin-adapter architecture, where a shallow adapter and a deep adapter progressively synthesize inter-modal complementary prompts.
        }
	\label{fig:arch}
	\vspace{-3mm}
\end{figure*}

\section{Methodology}
\label{sec:methodology}
In this section, DMTrack is presented step by step. First, we formulate the pipeline of video-level multimodal tracking. Next, we present an STMA designed for inner-modal spatio-temporal context self-prompting, followed by the introduction of a PMCA module that progressively generates cross-modal prompts on a pixel-wise basis. Finally, we introduce the prediction head and training objective function.

\subsection{Video-Level Multi-modal Tracking}
In contrast to image-level paradigms that rely on a single template image and a single search image as input, we construct a template memory bank $M \in \mathbb{R}^{T \times 3 \times H_z \times W_z}$ using historical frames. This memory bank, combined with a search frame $X \in \mathbb{R}^{3 \times H_x \times W_x}$, forms our input, thereby lifting the foundation model to the video level. As illustrated in Fig.~\ref{fig:arch}, our framework processes dual-modality video streams $\{Z_{RGB}^1, Z_{RGB}^2, \ldots, Z_{RGB}^k, X_{RGB}\}$ and $\{Z_{XM}^1, Z_{XM}^2, \ldots, Z_{XM}^k, X_{XM}\}$, which are temporally synchronized and spatially aligned. The core operation within the frozen transformer layers of each modality branch can be formulated as follows:
\begin{equation}
    \begin{split}
        Y_{RGB} &= \textnormal{Attn}([Z_{RGB}^1, Z_{RGB}^2,..., Z_{RGB}^k, X_{RGB}]) \\
        Y_{XM} &= \textnormal{Attn}([Z_{XM}^1, Z_{XM}^2, ..., Z_{XM}^k, X_{XM}]) 
    \end{split}
    \label{eq:f1}
\end{equation}
where $XM$ denotes the X modality (Thermal, Event, and Depth). $k$ is the length of the memory bank. By employing a uniform interval sampling strategy for frame selection, our method enables robust temporal information modeling while maintaining a uniform number of sampled frames. We avoid temporal propagation when incorporating temporal context, as it may lead to overfitting given the limited scale of multimodal training data.
With such designs, we simplify the video-level tracking pipeline, significantly reducing memory consumption during training and demonstrating that the memory bank is sufficient to provide robust spatio-temporal cues. 

\subsection{Spatio-Temporal Modality Adapter}
\begin{figure}[t]
  \centering
  \setlength{\abovecaptionskip}{1pt}
  \includegraphics[width=0.65\linewidth]{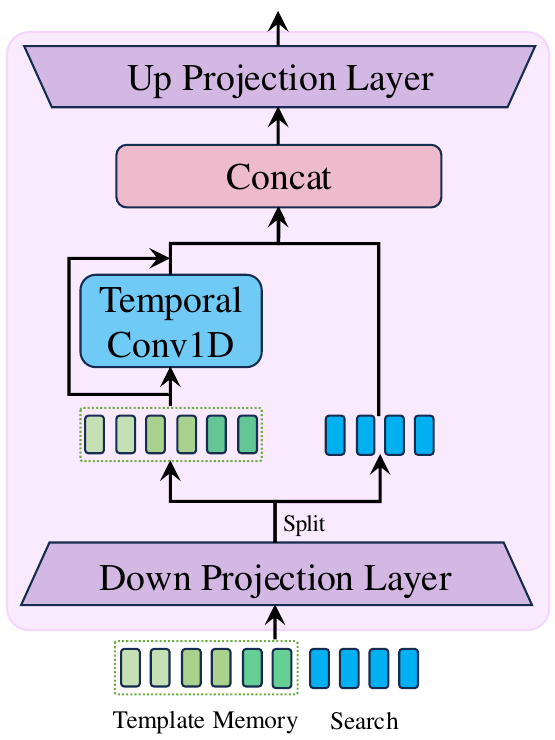}
\caption{Detailed design of STMA. In STMA, the temporal context is extracted from Template Memory via a 1D convolutional layer.
}
\vspace{-3mm}
\label{fig:ta}
\end{figure}

Previous spatio-temporal trackers have predominantly followed a brute-force paradigm, relying on global cross-modal interactions through full fine-tuning of entire networks. While such approaches achieve moderate performance given sufficient computational and parametric budgets, they suffer from inefficiency and suboptimal performance by neglecting the inherent modality gap between heterogeneous modality video streams. For instance, event video frames exhibit sparse spatio-temporal distribution due to their asynchronous triggering mechanism, while RGB video frames contain dense spatio-temporal variations with continuous photometric changes. To address this limitation, we propose an STMA that dynamically learns spatio-temporal cues for each modality branch with modality-specific parameters.  Designed in a modular fashion, STMA is integrated in the front of a transformer block, enabling parameter-efficient spatio-temporal self-prompting that reduces the gap between the two modalities in the high-dimensional feature space. As shown in Fig.~\ref{fig:ta}, for the input of each modality denoted as $X \in \mathbb{R}^{B \times N \times C}$, we split it into the search part and template part after the down-projection: 
\begin{equation}
    \begin{split}
        X_{\text{down}} &= XW_{\text{down}} + b_{\text{down}} \\
        X_x &= X_{\text{down}}[:, T \cdot N_x :] \\
        X_z &= X_{\text{down}}[:, :T \cdot N_z]
    \end{split}
\end{equation}

where $N_x$ and $N_z$ represent the length of search and template tokens, respectively. $T$ is the size of the template memory bank. After we reshape $X_z$ from $X_z \in \mathbb{R}^{B \times (N_z \cdot d) \times T}$ to $X_z \in \mathbb{R}^{(B \cdot N_z) \times d \times T}$, we perform the following operations: 
\begin{equation}
X_z' = X_z + \text{Conv1d}(X_z)
\end{equation}
where Conv1D denotes the 1D-convolution for spatio-temporal reasoning operating on the temporal dimension we introduce. It is noteworthy that after applying Conv1D, the $X_z'$ will be reshaped back from $X_z' \in \mathbb{R}^{(B \cdot N_z) \times d \times T}$ to $X_z' \in \mathbb{R}^{B \times (N_z \cdot d) \times T}$. Finally, $X_z'$ is concatenated with $X_x$ followed by the up-projection: 
\begin{equation}
    \begin{split}
        X_{\text{down}}' = \text{Concat}(X_z', X_x) \\
        X_{\text{up}} = X_{\text{down}}'W_{\text{up}}  + b_{\text{up}}
    \end{split}
\end{equation}
Consequently, the STMA enjoys high efficiency and effectiveness in spatio-temporal modeling while merely incorporating tiny extra (0.6\%) parameters.

\subsection{Progressive Modality Complementary Adapter}
The paradigm of generating complementary prompts for the other modality through pixel-wise operations has demonstrated promising results~\cite{vipt,bat}. Unlike BAT, which applies an identical processing strategy after both the MHA (multi-head attention) and MLP components within each ViT block, our proposed PMCA explicitly considers the difference in information density between these two stages. Leveraging this observation, PMCA introduces a progressive adaptation strategy composed of two complementary components: a shallow adapter and a deep adapter. Specifically, we adopt bi-directional adapter from BAT as our shallow adapter, which establishes inter-modal connectivity via parameter-shared transformations, creating a foundational feature bridge between each modality branch. On top of this, the deep adapter refines the fused features through dual pixel-wise attention mechanisms: intra-modal attention for feature recalibration and inter-modal attention for modality-aware prompting to guide cross-modal adaptation. 

\begin{figure}[t]
  \centering
  \setlength{\abovecaptionskip}{0.5pt}
  \includegraphics[width=0.57\linewidth]{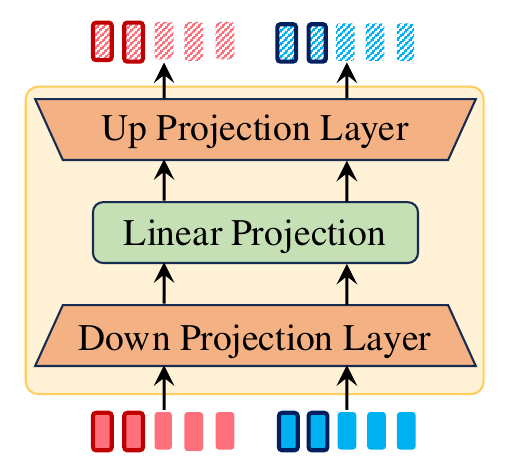}
  \caption{Detailed design of Shallow Adapter. Multimodal input flows are processed through three FC layers to generate foundational cross-modal complementary prompts, which are subsequently supplied to another modality branch.
  }
  \vspace{-3mm}      
  \label{fig:ca}
\end{figure}

\noindent \textbf{Shallow Adapter.} As illustrated in Fig.~\ref{fig:ca}, the shallow adapter includes a down-projection fully connected (FC) layer, an up-projection FC layer, and a linear FC layer. Formally, the shallow adapter can be expressed as:
\begin{equation}
    \begin{split}
    Y_{RGB \to X} &= ((X_{RGB}W_{\text{down}} )W_{\text{mid}})W_{\text{up}} \\
    Y_{X \to RGB} &= ((X_{X}W_{\text{down}})W_{\text{mid}})W_{\text{up}}
    \end{split}
\end{equation}
where $X_{RGB}$ and $X_{X}$ are the input tokens of RGB and X modality. Similar to the STMA, the shallow adapter employs a modular design and is integrated into the MHA stage. Since it serves as a foundational feature bridge between each modality branch, the weights are shared across different modality streams. Finally, the complementary information is merged into the other modality stream via element-wise addition. With such a simple but effective design, we establish initial cross-modal correspondences.

\begin{figure}[t]
  \centering
  \setlength{\abovecaptionskip}{0.5pt}
  \includegraphics[width=0.8\linewidth]{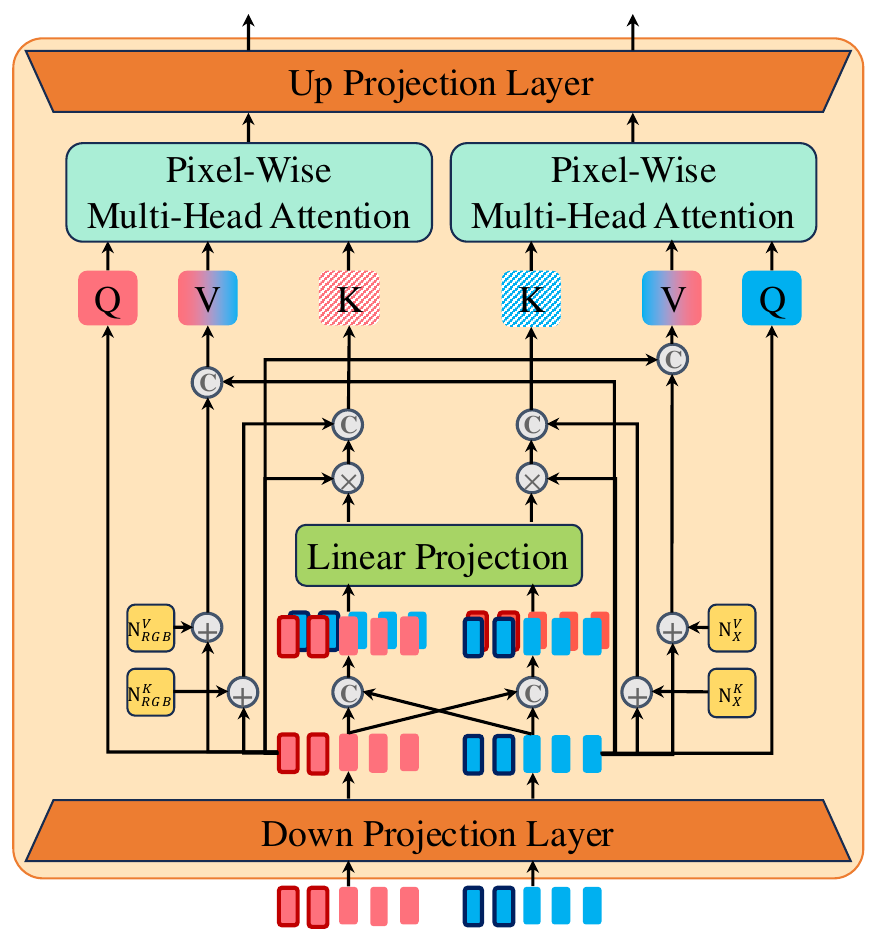}
        \caption{Detailed design of Deep Adapter. 
        In deep adapter, we construct both Key and Value using dual modalities, enabling pixel-wise attention to simultaneously refine intra-modal representations while adaptively fusing cross-modal information.
        }
	\label{fig:fa}
	\vspace{-3mm}
\end{figure}

\noindent \textbf{Deep Adapter.} Building upon the preliminary cross-modal interaction introduced by the shallow adapter, the deep adapter further leverages a pixel-wise MHA mechanism to generate modality-aware complementary prompts. As illustrated in Fig.~\ref{fig:fa}, given the input of RGB and X modality, \textit{i.e.}, $X_{RGB} \in \mathbb{R}^{B\times N\times C}$ and $ X_{X} \in \mathbb{R}^{B\times N\times C}$, we first project them to lower-dimensional of $d$. Considering the differences between modalities, we adopt a lightweight gating unit to compute relation-aware scores of X modality and RGB as:
\begin{equation}
    \begin{split}
        Score_{RGB \to X} &= \text{softmax}(\text{Concat}(X_{X}, X_{RGB})W_{gate}) \\
        Score_{X \to RGB} &= \text{softmax}(\text{Concat}(X_{RGB}, X_{X})W_{gate})
    \end{split}
\end{equation}
where $W_{gate}$ is the weight of the linear-projection. To prevent the bias introduced by query and key containing information from the same modality, we inject a layer-adaptive noise when computing the key and value as:
\begin{equation}
    \begin{split}
        Q_{RGB} &= X_{RGB} \\
        K_{RGB} &= \text{Concat}(X_{RGB} + N_{RGB}^k,\;  X_{RGB} \odot Score_{X \to RGB}) \\
        V_{RGB} &= \text{Concat}(X_{RGB} + N_{RGB}^v,\; X_{X}) \\
        Q_{X} &= X_{X} \\
        K_{X} &= \text{Concat}(X_{X} + N_{X}^k,\; X_{X} \odot Score_{RGB \to X}) \\
        V_{X} &= \text{Concat}(X_{X} + N_{X}^v,\; X_{RGB}), 
    \end{split}
    \label{eq:qkv}
\end{equation}

where $N_{X}^k$, $N_{X}^v$, $N_{RGB}^k$, $N_{RGB}^v$ are the learnable noise embeddings, and $\odot$ indicates the element-wise multiplication. As shown in Eq.~\ref{eq:qkv}, we integrate self-attention with cross-attention in the deep adapter. This dual mechanism simultaneously captures intra-modal dependencies and inter-modal interactions, thereby producing modality-aware complementary prompts. The attention is computed as:
\begin{equation}
    \begin{split}
        \text{P}_{RGB} &= \text{PW-MHA}(Q_{RGB},\;K_{RGB},\;V_{RGB}) \\
        \text{P}_{X} &= \text{PW-MHA}(Q_{X}, K_{X}, V_{X})
    \end{split}
\end{equation}

\noindent where $\text{PW-MHA}$ denotes the pixel-wise MHA. $\text{P}_{RGB}$ and $\text{P}_{X}$ represent the modality-aware complementary cues for the RGB and X branches, respectively. It is noteworthy that the core $\text{PW-MHA}$ mechanism employs modality-specific parameters. With such designs, we explicitly account for the complementarity of patches at corresponding spatial positions across different modalities. By employing a balanced combination of pixel-wise intra-modal self-attention and inter-modal cross-attention, we generate robust completion cues with minimal computational and parametric overhead. Finally, the $\text{P}_{RGB}$ and $\text{P}_{X}$ are projected back to the original dimension and merged into each modality stream via element-wise addition.

\subsection{Head and Objective Loss}
Following prevailing methodologies~\cite{ostrack} in visual tracking, our framework features a fully convolutional network-based prediction head. 
The overall loss function is
\begin{equation}
	\label{equ-loss-loc}
	\begin{aligned}
		\mathcal{L}=\mathcal{L}_{\operatorname{focal}} + \lambda_{G}\mathcal{L}_{\operatorname{GIoU}}+\lambda_{l}\mathcal{L}_l,
	\end{aligned}
\end{equation}
where $\lambda_{G}=2$ and $\lambda_{l}=5$ are the regularization parameters.

\section{Experiments}
\label{sec:experiments}
In this section, we first provide a detailed description of the experimental setup. Next, we compare DMTrack with other state-of-the-art (SOTA) methods across several benchmark datasets. Finally, the ablation study and qualitative comparison are presented.
\subsection{Implementation Details}
\noindent \textbf{Training.}
As a unified multimodal tracking framework, we present a versatile RGB-X tracker that flexibly addresses a range of tasks, including RGB-T, RGB-D, and RGB-E tracking. The training process leverages the LasHeR, DepthTrack, and VisEvent datasets. DMTrack is implemented in Python 3.8 using PyTorch 2.2.2 and trained on four NVIDIA RTX 3090 GPUs over 60 epochs, with each epoch comprising 60,000 sample pairs. The total batch size is set at 64. The search and template images are resized to 256 $\times$ 256 and 128 $\times$ 128, respectively. We employ AdamW~\cite{AdamW} optimizer with a weight decay of 1e-4 and initialize the learning rate at 4e-4, reducing it by 10\% during the final 20\% of the epochs.

\vspace{1mm}
\noindent \textbf{Inference.} In line with our training configuration, we integrate multiple template memory frames sampled at equal intervals into our tracker during inference. Evaluated on an NVIDIA RTX 3090 GPU, the tracker operates at approximately $39.21$ frames per second (FPS).

\begin{table*}[!t]
    \small
    \caption{
	\small
	Overall performance on DepthTrack test set~\cite{yan2021depthtrack}.
    }
    \vspace{-2.5mm}
    \centering
    \renewcommand\arraystretch{1.15} 
    \setlength{\tabcolsep}{1.5mm}{
    \resizebox{\linewidth}{!}{
	\begin{tabular}{c|cccccccccccc}
	\hline
	\small
	&\tabincell{c}{OSTrack\\~\cite{ostrack}}&\tabincell{c}{DeT\\~\cite{yan2021depthtrack}}&\tabincell{c}{SPT\\~\cite{spt}}&\tabincell{c}{ProTrack\\~\cite{protrack}}&\tabincell{c}{ViPT\\~\cite{vipt}}&\tabincell{c}{OneTracker\\~\cite{onetracker}}&\tabincell{c}{UnTrack\\~\cite{untrack}}&\tabincell{c}{SDSTrack\\~\cite{sdstrack}}&\tabincell{c}{SeqTrackv2\\~\cite{seqtrackv2}}&\tabincell{c}{STTrack\\~\cite{STTrack}}&\tabincell{c}{\textbf{DMTrack}\\\textbf{(Ours)}}\\
	\hline
	F-score($\uparrow$)&0.529&0.532&0.578&0.578&0.594&0.609&0.612&0.614&0.632&\textbf{\textcolor[rgb]{0,0,1}{0.633}}&\textbf{\textcolor[rgb]{1,0,0}{0.647}}\\
				Recall($\uparrow$)&0.522&0.506&0.538&0.573&0.596&0.604&0.610&0.609&0.634&\textbf{\textcolor[rgb]{0,0,1}{0.634}}&\textbf{\textcolor[rgb]{1,0,0}{0.648}}\\
				Precision($\uparrow$)&0.536&0.560&0.527&0.583&0.592&0.607&0.613&0.619&0.629&\textbf{\textcolor[rgb]{0,0,1}{0.632}}&\textbf{\textcolor[rgb]{1,0,0}{0.647}}\\
				\hline
			\end{tabular}
	}}\\
	\vspace{-2mm}
	\label{tab-depthtrack}
\end{table*}

\begin{table*}[!t]
    \small
    \caption{
		\small
		Overall performance on VOT-RGBD2022~\cite{votd22}.
    }
    \vspace{-2.5mm}
	\centering
	\renewcommand\arraystretch{1.15} 
	\setlength{\tabcolsep}{1.5mm}{ 
		\resizebox{\linewidth}{!}{
			\begin{tabular}{c|cccccccccccccc}
				\hline
				\small
				&\tabincell{c}{KeepTrack\\~\cite{keeptrack}}&\tabincell{c}{STARK-RGBD\\~\cite{stark}}&\tabincell{c}{SPT\\~\cite{spt}}&\tabincell{c}{ProTrack\\~\cite{protrack}}&\tabincell{c}{DeT\\~\cite{yan2021depthtrack}}&\tabincell{c}{OSTrack\\~\cite{ostrack}}&\tabincell{c}{SBT-RGBD\\~\cite{sbt}}&\tabincell{c}{ViPT\\~\cite{vipt}}&\tabincell{c}{UnTrack\\~\cite{untrack}}&\tabincell{c}{OneTracker\\~\cite{onetracker}}&\tabincell{c}{SDSTrack\\~\cite{sdstrack}}&\tabincell{c}{SeqTrackv2\\~\cite{seqtrackv2}}&\tabincell{c}{STTrack\\~\cite{STTrack}}&\tabincell{c}{\textbf{DMTrack}\\\textbf{(Ours)}}\\
				\hline
				EAO($\uparrow$)&0.606&0.647&0.651&0.651&0.657&0.676&0.708&0.721&0.718&0.727&0.728&0.744&\textbf{\textcolor{blue}{0.776}}&\textbf{\textcolor[rgb]{1,0,0}{0.794}} \\
				Accuracy($\uparrow$)&0.753&0.803&0.798&0.801&0.760&0.803&0.809&0.815&0.820&0.819&0.812&0.815&\textbf{\textcolor{blue}{0.825}}&\textbf{\textcolor[rgb]{1,0,0}{0.837}} \\
				Robustness($\uparrow$)&0.739&0.798&0.851&0.802&0.845&0.833&0.864&0.871&0.864&0.872&0.883&0.910&\textbf{\textcolor{blue}{0.937}}&\textbf{\textcolor[rgb]{1,0,0}{0.943}} \\
				\hline
			\end{tabular}
	}}
	\vspace{-3mm}
	\label{tab-vot22rgbd}
\end{table*}
\begin{table*}[!t]
    \small
    \caption{
		\small
		Overall performance on VisEvent~\cite{wang2023visevent} test set.
    }
    \vspace{-2.5mm}
	\centering
	\renewcommand\arraystretch{1.15} 
	\setlength{\tabcolsep}{1.5mm}{ 
		\resizebox{\linewidth}{!}{
			\begin{tabular}{c|cccccccccccc}
				\hline
				\small
                    &\tabincell{c}{LTMU\_E\\~\cite{ltmu}}&\tabincell{c}{ProTrack\\~\cite{protrack}}&\tabincell{c}{TransT\_E\\~\cite{transt}}&\tabincell{c}{SiamRCNN\_E\\~\cite{siamrcnn}}&\tabincell{c}{OSTrack\\~\cite{ostrack}}&\tabincell{c}{UnTrack\\~\cite{untrack}}&\tabincell{c}{ViPT\\~\cite{vipt}}&\tabincell{c}{SDSTrack\\~\cite{sdstrack}}&\tabincell{c}{OneTrack\\~\cite{onetracker}}&\tabincell{c}{SeqTrackV2\\~\cite{seqtrackv2}}&\tabincell{c}{STTrack\\~\cite{STTrack}}&\tabincell{c}{\textbf{DMTrack}\\\textbf{(Ours)}} \\ 
				\hline
				AUC($\uparrow$)&45.9 &47.1 &47.4 &49.9 &53.4 &58.9 &59.2 &59.7 &60.8 &61.2 &\textbf{\textcolor{blue}{61.9}} &\textbf{\textcolor[rgb]{1,0,0}{62.4}}\\
                    PR($\uparrow$)&65.9 &63.2 &65.0 &65.9 &69.5 &75.5 &75.8 &76.7 &76.7 &78.2 &\textbf{\textcolor{blue}{78.6}} &\textbf{\textcolor[rgb]{1,0,0}{79.6}}\\                
				\hline
			\end{tabular}
	}}
	\vspace{-3mm}
	\label{tab-sota-rgbe}
\end{table*}

\begin{table*}[!t]
    \caption{
	\small
	Overall performance on LasHeR\cite{li2021lasher} test set.
    }
    \vspace{-2.5mm}
	\small
	\centering
	\renewcommand\arraystretch{1.15} 
	\setlength{\tabcolsep}{1.5mm}{ 
		\resizebox{\linewidth}{!}{
			\begin{tabular}{c|cccccccccccccc}
				\hline
				\small
				&\tabincell{c}{ProTrack\\~\cite{protrack}}&\tabincell{c}{OSTrack\\~\cite{ostrack}}&\tabincell{c}{ViPT\\~\cite{vipt}}&\tabincell{c}{SDSTrack\\~\cite{sdstrack}}&\tabincell{c}{UnTrack\\~\cite{untrack}}&\tabincell{c}{OneTracker\\~\cite{onetracker}}&\tabincell{c}{CAFormer\\~\cite{caformer}}&\tabincell{c}{SeqTrackv2\\~\cite{seqtrackv2}}&\tabincell{c}{TATrack\\~\cite{tatrack}}&\tabincell{c}{TBSI\\~\cite{tbsi}}&\tabincell{c}{BAT\\~\cite{bat}}&\tabincell{c}{GMMT\\~\cite{gmmt}}&\tabincell{c}{STTrack\\~\cite{STTrack}}&\tabincell{c}{\textbf{DMTrack}\\\textbf{(Ours)}}\\
				\hline
                    PR($\uparrow$)&53.8&52.5&65.1&66.5&66.7&67.2&70.0&70.4&70.2&70.5&70.2&70.7&\textbf{\textcolor{blue}{76.0}}&\textbf{\textcolor[rgb]{1,0,0}{76.1}} \\
                    SR($\uparrow$)&42.0&41.2&52.5&53.1&53.6&53.8&55.6&55.8&56.1&56.3&56.3&56.6&\textbf{\textcolor{blue}{60.3}}&\textbf{\textcolor[rgb]{1,0,0}{60.3}} \\
				\hline
			\end{tabular}
	}}
	\vspace{-3mm}
	\label{tab-lasher}
\end{table*}

\begin{table*}[!t]
    \small
    \caption{
		\small
		Overall performance on RGBT234\cite{rgbt234}.
        }
        \vspace{-2.5mm}
	\centering
	\renewcommand\arraystretch{1.15} 
	\setlength{\tabcolsep}{1.5mm}{ 
		\resizebox{\linewidth}{!}{
			\begin{tabular}{c|cccccccccccccc}
				\hline
				\small
				&\tabincell{c}{ProTrack\\~\cite{protrack}}&\tabincell{c}{OSTrack\\~\cite{ostrack}}&\tabincell{c}{ViPT\\~\cite{vipt}}&\tabincell{c}{SDSTrack\\~\cite{sdstrack}}&\tabincell{c}{UnTrack\\~\cite{untrack}}&\tabincell{c}{OneTracker\\~\cite{onetracker}}&\tabincell{c}{CAFormer\\~\cite{caformer}}&\tabincell{c}{SeqTrackv2\\~\cite{seqtrackv2}}&\tabincell{c}{TATrack\\~\cite{tatrack}}&\tabincell{c}{TBSI\\~\cite{tbsi}}&\tabincell{c}{BAT\\~\cite{bat}}&\tabincell{c}{GMMT\\~\cite{gmmt}}&\tabincell{c}{STTrack\\~\cite{STTrack}}&\tabincell{c}{\textbf{DMTrack}\\\textbf{(Ours)}}\\
				\hline
				MPR($\uparrow$)&79.5&72.9&83.5&84.8&83.7&85.7&88.3&88.0&87.2&86.4&86.8&87.9&\textbf{\textcolor{blue}{89.8}}&\textbf{\textcolor[rgb]{1,0,0}{90.3}} \\
                    MSR($\uparrow$)&59.9&54.9&61.7&62.5&61.8&64.2&66.4&64.7&64.4&64.3&64.1&64.7&\textbf{\textcolor[rgb]{1,0,0}{66.7}}&\textbf{\textcolor{blue}{65.7}} \\
				\hline
			\end{tabular}
	}}
	\vspace{-5mm}	
	\label{tab-rgbt234}
\end{table*}

\subsection{Comparison with State-of-the-Arts}
\noindent \textbf{DepthTrack}. DepthTrack is a long-term RGB-D tracking benchmark with an average sequence length of 1,473 frames. It includes 200 sequences across 40 scenes and 90 target objects. As shown in Table.~\ref{tab-depthtrack}, our DMTrack achieves SOTA results, with an F-score of 64.7\%, recall of 64.8\%, and precision of 64.7\%.

\vspace{1mm}
\noindent \textbf{VOT-RGBD2022}. VOT-RGBD2022 consists of 127 short-term RGB-D sequences and evaluates tracker performance with Accuracy, Robustness, and Expected Average Overlap (EAO). As demonstrated in Table.~\ref{tab-vot22rgbd}, DMTrack achieves an EAO score of 79.4\%, accuracy of 83.7\%, and robustness of 94.3\%, outperforming the previous SOTA tracker STTrack.

\vspace{1mm}
\noindent \textbf{VisEvent}. VisEvent, as a large-scale RGB-E dataset, comprises 500 training and 320 testing video sequences. As reported in Table.~\ref{tab-sota-rgbe}, our DMTrack achieves SOTA performance with an AUC of 62.4\% and a precision of 79.6\%.

\vspace{1mm}
\noindent \textbf{LasHeR}. LasHeR is a large-scale RGB-T tracking benchmark, consisting of 1,224 aligned sequences. As shown in Table.~\ref{tab-lasher}, our DMTrack achieves a success rate (SR) of 60.3\% and a precision rate (PR) of 76.1\%, outperforming the previous SOTA tracker STTrack by 0.1\% in PR. This highlights the effectiveness of continuous spatio-temporal thermal information modeling of DMTrack.

\vspace{1mm}
\noindent \textbf{RGBT234}. RGBT234, extended from RGBT210~\cite{rgbt210} dataset, incorporates a broader range of environmental challenges, consisting of 234 aligned RGBT sequences. As shown in Table.~\ref{tab-rgbt234}, DMTrack achieves the highest MPR score of 90.3\%, exhibiting very competitive performance.

\subsection{Ablation Study}
\begin{table}[!t]
\caption{Ablation of various components. Each row is the baseline minus some DMTrack component. `$\Delta$' denotes the averaged performance change.
}
\vspace{-2.5mm}
\centering
\small
\fontsize{7}{9}\selectfont
\begin{tabular}{l|ccc|c}
\toprule
\multicolumn{1}{c|}{Model Variants} & LasHeR & Visevent & DepthTrack & $\Delta$ \\
\midrule 
DMTrack & 60.3 & 62.4 & 64.7 & - \\
w/o STMA & 58.7 & 62.0 & 64.5 & -0.73 \\
w/o STMA \& Memory Bank & 56.5 & 60.3 & 61.4 & -3.07 \\
w/o Shallow Adapter & 59.5 & 62.1 & 62.4 & -1.13 \\   
w/o Deep Adapter & 58.5 & 62.3 & 62.6 & -1.33 \\ 
\bottomrule
\end{tabular}
\vspace{-7mm}
\label{tab:ab_1}
\end{table}
\noindent \textbf{Component Analysis}.
In Table.~\ref{tab:ab_1}, comprehensive ablation studies are conducted to analyze key components of DMTrack. We select AUC in LasHeR, PR in DepthTrack, and AUC in VisEvent as the evaluation metrics. From the results, we observed that the incorporation of temporal information is the most critical factor for performance improvements. When both the memory bank and STMA are removed from the model, DMTrack is reduced to a non-temporal tracker, resulting in the most significant performance degradation. The incorporation of STMA, built upon the memory bank, yields substantial benefits, which demonstrates its ability to facilitate the model in learning the appearance evolution of the target in the memory bank. 
Additionally, the results reveal that either the absence of basic modality complementary prompts (resulting in blocked bidirectional information flow) or the lack of modality-aware complementary prompts leads to severe performance degradation.

\begin{figure*}[!t] 
  \centering
    \includegraphics[width=0.95\linewidth] {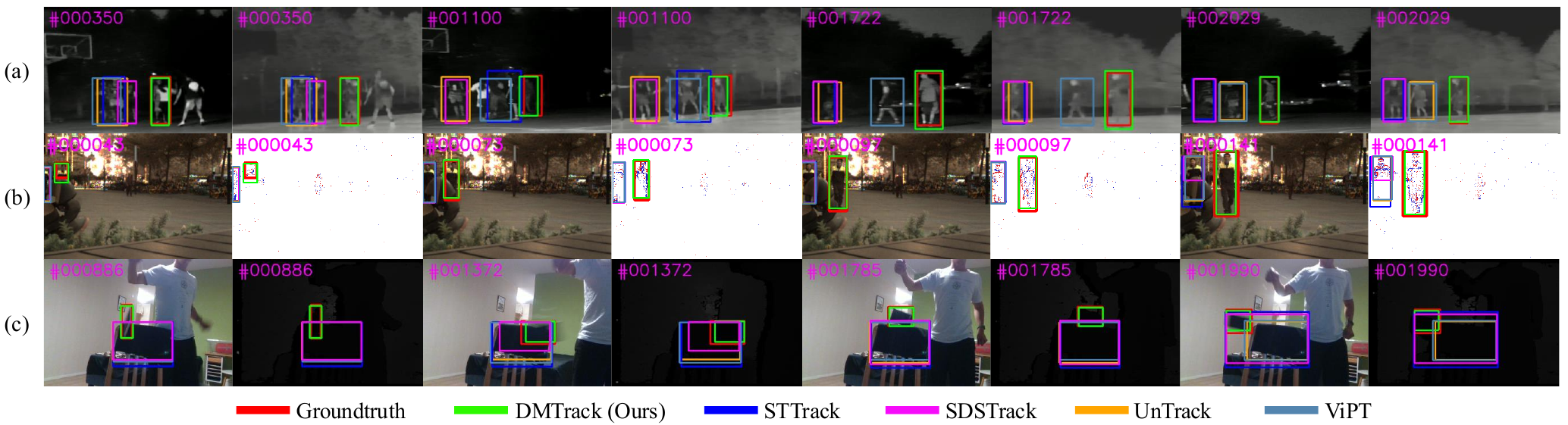}
    \vspace{-4mm}
    \caption{
    Qualitative comparison with SOTA unified multimodal trackers across three challenging scenarios: (a) nighttime crowded environments, (b) severe occlusion, and (c) similar distractors. DMTrack demonstrates accuracy and temporal consistency via effective spatio-temporal modeling capabilities.
    }
   \label{fig:qualitative_results}
   \vspace{-5mm}
\end{figure*}

\setlength{\subfigcapskip}{-9.5pt}
\begin{figure*}[!t]
  \centering
  \setlength{\abovecaptionskip}{1pt}
  \subfigure[Attributes within VisEvent.]{
    \includegraphics[width=0.45\linewidth]{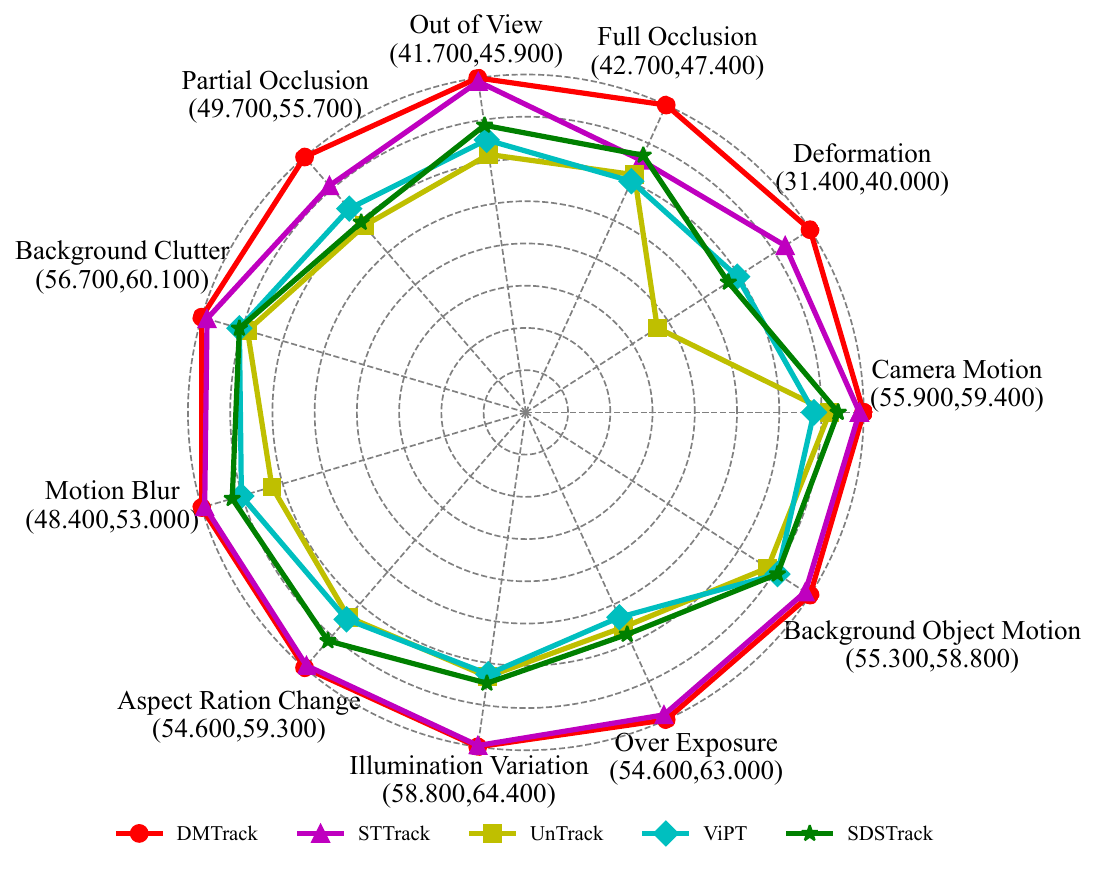}
    \label{fig:attr_e}
  }
  \hfill
  \subfigure[Attributes within LasHeR.]{
    \includegraphics[width=0.45\linewidth]{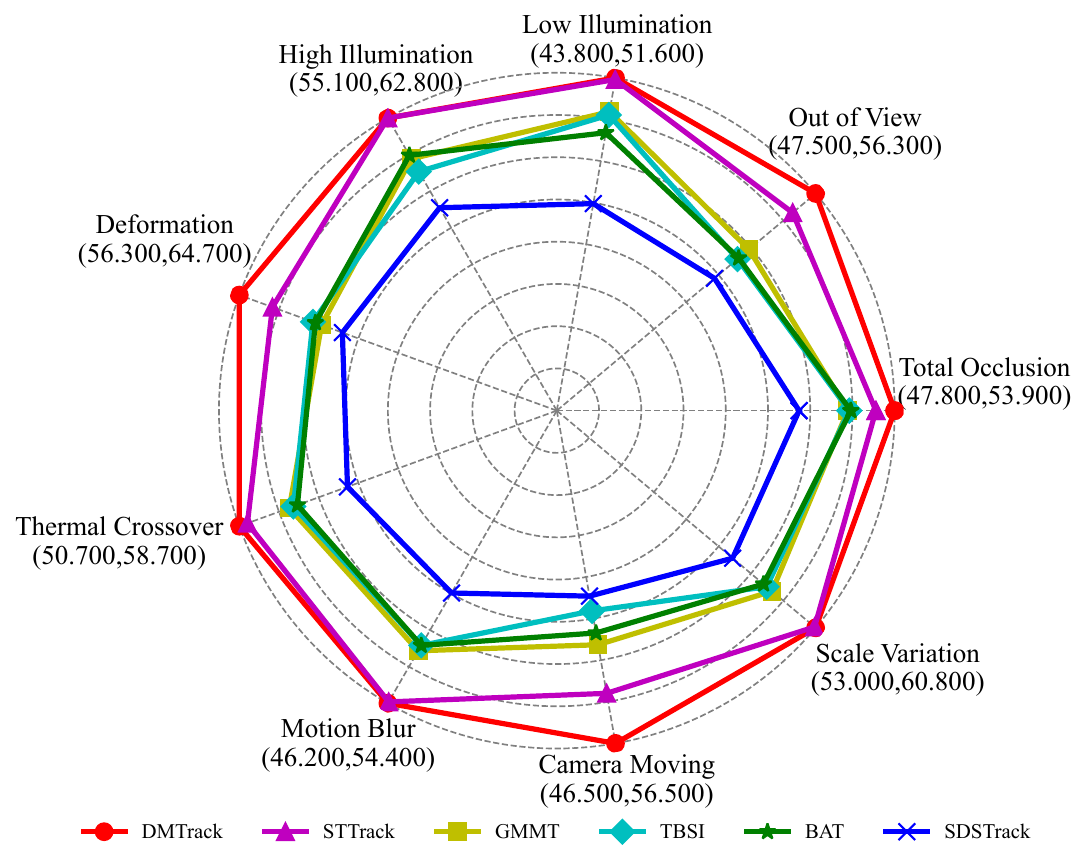}
    \label{fig:attr_t}
  }
  \vspace{-1mm}
  \caption{Comprehensive comparison between DMTrack and SOTA trackers under challenging attributes within VisEvent (a) and LasHeR (b).}
  \vspace{-5mm}
\end{figure*}

\begin{table}[!t]\normalsize
\centering
\caption{Ablation study on the size of the template memory bank. \colorbox{gray!18}{Gray} denotes our final configuration.}
\vspace{-2.5mm}
\fontsize{6}{6}\selectfont
    \small
    \begin{tabular}{c|ccc}
    \toprule
    Memory size & LasHeR & VisEvent & DepthTrack\\
    \midrule
    2 &59.4 &61.7 &\cellcolor{gray!15}64.7\\
    3 &\cellcolor{gray!15}60.3  &\cellcolor{gray!15}62.4 &63.0\\
    4 &60.0 & 62.2 & 64.4\\
    \bottomrule
    \end{tabular}
    \vspace{-4mm}
    \label{tab:ab_2}
\end{table}

\vspace{1mm}
\noindent \textbf{Memory Bank Size}.
In DMTrack, a key design is the incorporation of a memory bank comprised of historical frames. The historical states provide critical cues of target changes and motion trajectories. The memory bank size represents the length of the temporal information we maintain. In multimodal tasks, different modalities exhibit varying sensitivities to temporal information. Excessive temporal information can introduce disruptive noise, increasing the learning burden for the model. Therefore, as shown in Table.~\ref{tab:ab_2}, we explore the optimal memory bank size for each modality.

\begin{table}[!t]
\centering
\small
\caption{Ablation study on hidden states size and modality sharing in STMA. \colorbox{gray!18}{Gray} denotes our final configuration.}
\vspace{-4mm}
\fontsize{7}{8}\selectfont
\begin{tabular}{c|ccc}
\toprule
\multicolumn{1}{c|}{Method} & LasHeR & Visevent & DepthTrack \\
\midrule 
DMTrack & 60.3 & 62.4 & 64.7  \\
Modality Shared & 59.0 & 62.2 & 62.0  \\
\hline
8 hidden states & 60.0 & 62.1 & 64.6 \\
12 hidden states & 59.9 & \default{\textbf{62.4}} & \default{\textbf{64.7}} \\ 
16 hidden states & \default{\textbf{60.3}} & 62.0  & 62.5 \\
\bottomrule
\end{tabular}
\vspace{-7mm}
\label{tab:ab_3}
\end{table}

\vspace{1mm}
\noindent \textbf{Ablation of STMA}.
STMA is a critical component of DMTrack, responsible for facilitating the capture of inner-modal spatio-temporal cues. Therefore, we conduct ablation studies on whether parameters are shared across modalities and the size of the hidden states. The results are presented in Table.~\ref{tab:ab_3}. We found that when the spatio-temporal information across the two modality video streams is modeled using shared parameters, performance significantly degrades. This supports our hypothesis that video streams from different modalities exhibit distinctly different spatio-temporal information densities, and thus, separate parameters should be employed. We further investigated the optimal hidden state size of STMA for every modality. 

\vspace{-1mm}
\subsection{Visualization and Analysis}
\noindent \textbf{Qualitative Comparison.}
To further show tracking performance, we qualitatively compare DMTrack with four other SOTA multimodal trackers in Fig.~\ref{fig:qualitative_results}. Leveraging historical memory and progressive cross-modal prompts, DMTrack addresses a range of challenges such as motion blur and severe occlusion, thereby achieving robust tracking performance.

\vspace{1mm}
\noindent \textbf{Attribute-based Performance.}
Leveraging the rich attribute annotations provided by the VisEvent and LasHeR datasets, we select multiple representative attributes from each benchmark to analyze the performance of our method across various scenarios. As depicted in Fig.~\ref{fig:attr_t} and Fig.~\ref{fig:attr_e}, DMTrack outperforms previous SOTA trackers on all attributes. 
The results compellingly demonstrate that our method achieves exceptional robustness across a wide range of challenging scenarios.

\vspace{-2mm}
\section{Conclusion}
\label{sec:conclusion}
In this work, we present DMTrack, a parameter-efficient spatio-temporal tracking framework that incorporates two novel components: (1) The Spatio-Temporal Modality Adapter, which dynamically self-prompts modality-specific spatio-temporal cues through lightweight history template adaptation, and (2) The Progressive Modality Complementary Adapter module, which facilitates progressive cross-modal prompting via efficient pixel-wise operations. Experiments show that DMTrack is highly effective, achieving SOTA performance across multiple datasets. 

\section*{ACKNOWLEDGMENT}
Libo Zhang is supported by National Natural Science Foundation of China (No. 62476266). Heng Fan is not supported by any fund for this work.






\bibliographystyle{IEEEtran}
\bibliography{main}

\end{document}